\documentclass[conference]{IEEEtran}
\IEEEoverridecommandlockouts

\usepackage{cite}
\usepackage{amsmath,amssymb,amsfonts}
\usepackage{algorithmic}
\usepackage{graphicx}
\usepackage{textcomp}
\usepackage{xcolor}
\usepackage{multirow}
\usepackage{booktabs}
\usepackage{url}
\usepackage{color}
\usepackage[T1]{fontenc}
\usepackage{pifont}

\usepackage{makecell}

\def\BibTeX{{\rm B\kern-.05em{\sc i\kern-.025em b}\kern-.08em
    T\kern-.1667em\lower.7ex\hbox{E}\kern-.125emX}}
\begin{document}
\bstctlcite{BSTcontrol}

\title{VideoRun2D Demo: Markerless Body Tracking \\for Biomechanical Analysis of Running}

\author{

\IEEEauthorblockN{ Luis F. Gomez\IEEEauthorrefmark{1}, Julian Fierrez\IEEEauthorrefmark{1}, Roberto Daza\IEEEauthorrefmark{1}, Ruben Tolosana\IEEEauthorrefmark{1}, Aythami Morales\IEEEauthorrefmark{1}, }

\IEEEauthorblockN{Gonzalo Garrido\IEEEauthorrefmark{2}, Javier Rueda\IEEEauthorrefmark{2}, Enrique Navarro\IEEEauthorrefmark{2}} 

\vspace{.2cm}

\IEEEauthorblockA{\IEEEauthorrefmark{1}BiometricsAI, Universidad Autonoma de Madrid (UAM), Spain}

\IEEEauthorblockA{\IEEEauthorrefmark{2}Biomechanics Lab, Universidad Politecnica de Madrid (UPM), Spain}

{\tt julian.fierrez@uam.es, enrique.navarro@upm.es}
%
}

\maketitle

\begin{abstract}
Human pose estimation has advanced significantly due to the development of deep learning models, increased data availability, and improved computing resources. These developments have led to highly accurate body tracking systems with direct applications in sports analysis and performance evaluation.
The VideoRun2D Demo performs a biomechanical analysis during sprints using different human pose estimators. The proposed framework was evaluated using human pose trackers and expert manual annotations.
The tested framework uses 314 sprints from 44 professional runners, focusing on two key joint angles in sprint biomechanics: 1) hip flexion/extension and 2) knee flexion/extension. The framework also includes a post-processing module for outlier detection.
The tested results demonstrate that the average root-mean-square errors range from 11.46° to 5.83° for the best trackers. When integrated with the post-processing modules, these errors can be reduced to 9.87° and 5.30°, respectively. The VideoRun2D Demo findings suggest that human pose-tracking approaches can be valuable resources for the biomechanical analysis of running. 
\end{abstract} 


\section{Introduction}\label{sec:Introduction}

Human tracking technologies have advanced rapidly, driven by advances in machine learning and sensor fusion, creating new opportunities for motion analysis~\cite{daza2024improveimpactmobilephones}. In sports science, these advances enable for a more accurate evaluation of athletic performance, support injury prevention, and facilitate personalized training strategies. Tracking technologies also provide real-time feedback, helping athletes and coaches refine technique and improve performance during training.

Sprinting plays a crucial role in numerous sports and is a key factor in athletic performance~\cite{morin2011technical}. Therefore, it is essential to understand sprint mechanics to improve performance and prevent injuries. Historically, movement evaluation has relied on expert visual evaluation, which is inherently subjective and depends on the expertise of the evaluator~\cite{hii2023automated,yang2024improving}. To overcome these limitations, manual annotation was introduced to provide objective metrics for the analysis of timing parameters and joint kinematics~\cite{bissas2022kinematic,hanley2022biomechanics}.

Technological progress has also enabled automated motion analysis tools. Among these, marker-based systems provide high accuracy, but require controlled settings, specialized staff, and costly equipment~\cite{paula2023gait, viswakumar2022development, yang2024improving}. Inertial Measurement Units, on the contrary, present limitations such as sensor drift and questionable accuracy in joint-angle estimation~\cite{acien2020sensors,bastiaansen2020inertial,lin2023validity,nazarahari2022foot}.

As a more accessible alternative, smartphone applications have also been proposed for sprint analysis. However, these tools generally do not provide detailed joint kinematics, which limits their biomechanical utility~\cite{romero2017sprint}. 
Markerless motion capture has emerged as another promising alternative, leveraging deep learning-based estimators to track motion directly from video. These systems offer a balance between cost and analytical capability, making them attractive for sports applications. However, they still face challenges related to computational cost, the need for human supervision, and limited biomechanical accuracy~\cite{stenum2021two,yang2024improving}.

In the VideoRun2D Demo, we analyze sprint mechanics using markerless motion capture and assess the potential of the application of deep learning-based to enhance biomechanical analysis. VideoRun2D Demo evaluates state-of-the-art 2D pose estimation frameworks for sprint biomechanics, incorporating advanced signal post-processing techniques to improve kinematic accuracy. The main contributions of this work are as follows.

\begin{itemize}
    \item We present VideoRun2D Demo, a low-cost, non-invasive alternative for the biomechanical analysis of runners, based on previous developments\cite{garrido2024videorun2d}.
    \item We evaluated five approaches to human pose estimation against manually annotated ground truth to quantify biomechanical errors across configurations.
    \item We develop a fusion-tracking strategy that combines multiple trackers to enhance spatial robustness.
    \item We introduce an interactive web platform as a demonstrator.\footnote{\url{https://videorun2d.com/}} This platform introduces multiple trackers and supports different biomechanical analyzes for sprinters.
\end{itemize}

These contributions aim to enable cost-effective and accessible sprint analysis pipelines, with potential impact on professional sports performance and beyond, e.g. realistic computer animation \cite{ivan2025visualrealism}, biometric identification and re-identification \cite{nguyen25reid} at a distance \cite{2013_TIFS_PTome_SoftBiometrics} based on gait information as a soft biometric \cite{2018_TIFS_SoftWildAnno_Sosa} complementing other traditional biometrics that may not be fully discriminative in some scenarios like forensics \cite{2013PTomeFSI_FacialRegions } and surveillance \cite{cava253D,cava2025face3d}, and e-health applications \cite{ROMEROTAPIADOR2026111676,idea26protocol} based on gait information \cite{idea2023gait,idea24gait}.

\begin{figure*}
    \centering
    \includegraphics[width=\textwidth]{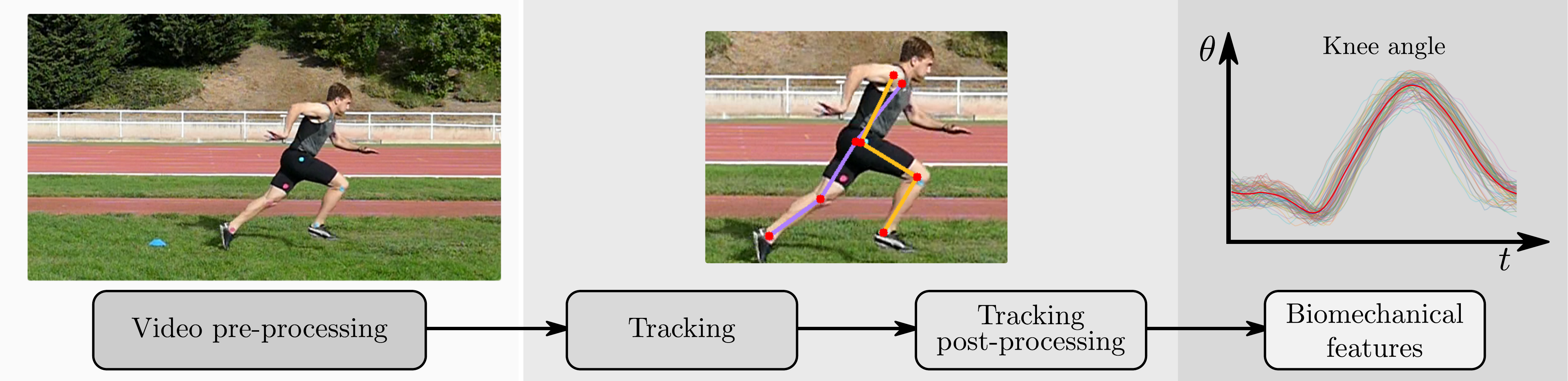}
    \caption{Block diagram of the VideoRun2D system. The system has five modules that estimate joint angles. }
    \label{fig:BlockDiagram}
\end{figure*}

\subsection{Related Work}

The use of markerless systems for sprint biomechanics has recently attracted increasing attention. Early attempts, e.g., using Kinect, showed promise, but suffered limited accuracy in joint-angle estimation, particularly in low-light conditions and certain clothing~\cite{ota2021verification,takeichi2018mobile,viswakumar2022development}.

More recently, advances in pose estimation have enabled the tracking of anatomical keypoints directly from video using deep networks~\cite{acien2020sensors, viswakumar2022development, yang2024improving}. Although these methods perform well in visual tracking, their biomechanical validity remains under discussion due to errors in joint-angle estimation.

Previous studies have mainly focused on keypoint detection accuracy or joint-angle estimation under controlled conditions. In 2022,~\cite{chung2022comparative} and~\cite{jo2022comparative} evaluated 4 pose estimation models in sports activities, but only assessed keypoint detection performance, reporting precisions between 70\% and 80\%. In 2024,~\cite{yang2024improving} evaluated the angles of the hip, knee, and ankle joint using a multi-camera markerless setup and reported errors below 5° relative to a marker system. Similarly,~\cite{galasso2024novel} found 93\% agreement between OpenPose~\cite{cao2019openpose} and an IMU system.

Last year,~\cite{gomez2025comparison} compared visual tracking models to perform biomechanics, including joint pose estimators and point-based tracking. Joint-based models achieved RMSEs between 11.4° and 4.4°, which were reduced to 7.0° and 3.9° after post-processing, highlighting the importance of refinement techniques for biomechanical applications.

Despite this progress, angular errors above 5° still limit clinical interpretability~\cite{lee2009running,mcginley2009reliability}. Moreover, only a limited number of studies have examined whether pose estimation models provide sufficient precision for sprint joint kinematic analysis. The present study addresses this gap by systematically investigating the biomechanical validity of different pose estimation frameworks in sprint-specific scenarios.



\section{VideoRun2D: System}\label{sec:VideoRunSystem}

VideoRun2D performs markerless body tracking and estimates joint angles during a sprint (see Figure~\ref{fig:BlockDiagram}). 
%
%

\subsection{VideoRun2D: Modules}

\subsubsection{Video pre-processing}

Our pre-processing module comprises a multiple-person detector, with a primary focus on the nearest person in the scene. After detection, a squared crop is used, reducing the input image from $1920\times1080$ to $300\times300$.

\subsubsection{Tracking} \label{sec:Tracking}

We evaluated five 2D human pose estimators to track the shoulder, elbow, hand, hip, knee, and ankle joints on both sides of the body. The models evaluated are ViTPose-B, ViTPose-L~\cite{xu2022vitpose}, RTMPose-M, and two versions of RTMPose-L~\cite{jiang2023rtmpose}.

\begin{itemize}

\item ViTPose, proposed by Xu \textit{et al.} in~\cite{xu2022vitpose}, uses Vision Transformers (ViTs) for human pose estimation instead of conventional convolutional backbones. In this study, ViTPose-B and ViTPose-L correspond to the ViT-Base and ViT-Large variants, with 86M and 307M parameters, respectively.

\item RTMPose, proposed in~\cite{jiang2023rtmpose} and implemented through OpenMMLab's MMPose~\cite{mmpose2020}, is based on a hybrid CNN-transformer backbone. In this work, we use RTMPose-M and RTMPose-L with an input size of $256\times192$, and RTMPose-L with an input size of $384\times288$.

\end{itemize}

\subsubsection{Tracking post-processing}\label{sec:PP_TrackingModule}

After pose trackers, three types of error were identified in knee and ankle joints due to occlusions and sagittal-plane ambiguities: 1) missing points, 2) confusion between the left and right and 3) misallocation of body-points.

To correct left-right confusion, the tracked trajectories $x_j(t)$ and $y_j(t)$ are smoothed using SVR, producing $\widetilde{x}_j(t)$ and $\widetilde{y}_j(t)$. The error curve is computed as $e_{m,j}(t)=|m_j(t)-\widetilde{m}_j(t)|2$, where $m_j(t)$ denotes either coordinate trajectory. Outliers are defined as samples that satisfy $e_{m,j}(t)>3\sigma_e$, where $\sigma_e$ is the standard deviation of the error distribution. Each outlier is first corrected by swapping the left and right assignments; otherwise, it is replaced by the smoothed SVR estimate $m_j(t_{\mathrm{out}})=\widetilde{m}_j(t_{\mathrm{out}})$.

\begin{figure}
    \centering
    \includegraphics[width=0.95\columnwidth]{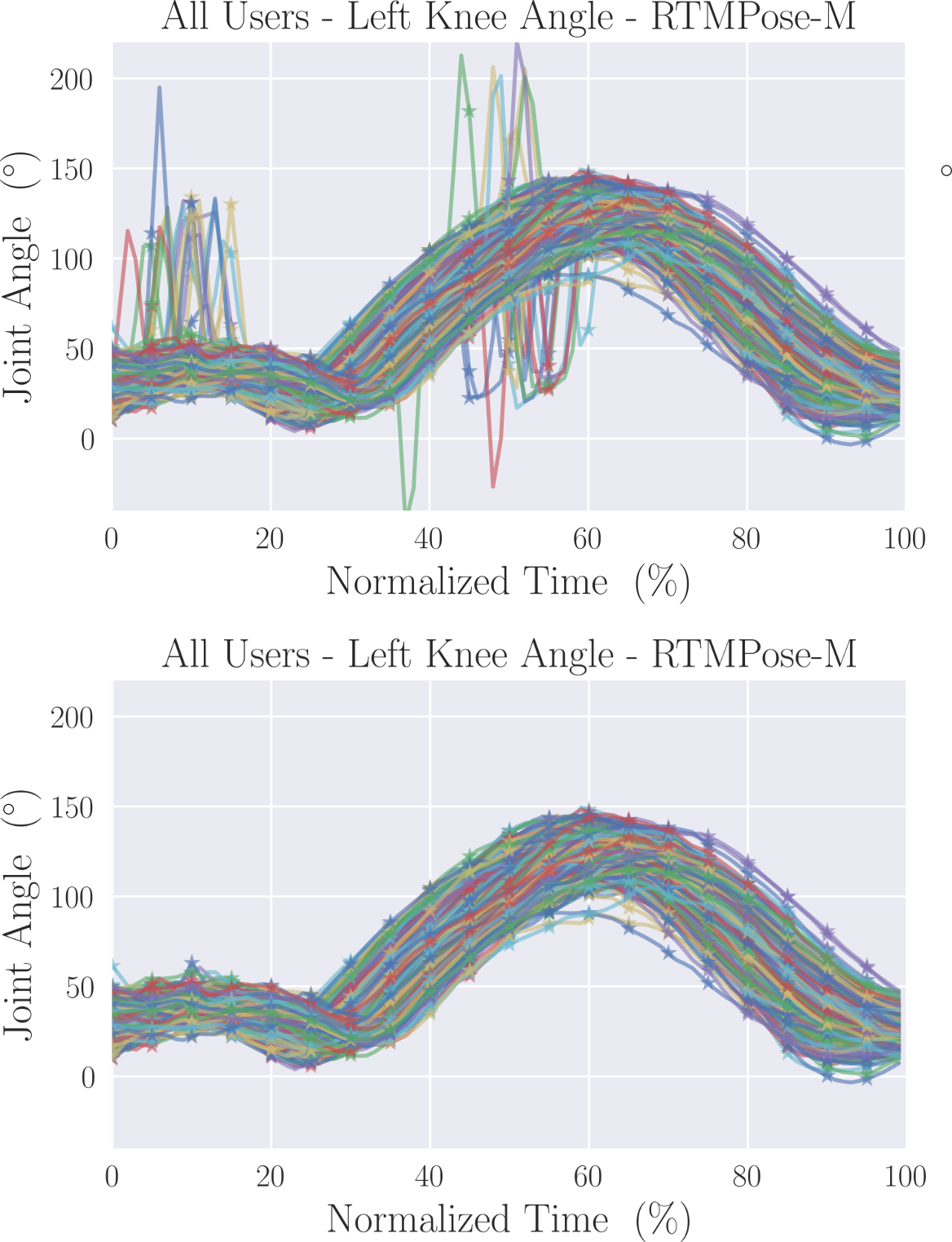}
    \caption{Graphical example of the correction of the left knee angle in all the thirteen strides from the sprinter two, both before (on the left) and after (on the right) post-processing}
    \label{fig:PP_UserStride}
\end{figure}

Figure~\ref{fig:PP_UserStride} shows a tracking result before and after the proposed post-processing for the angle of the left knee of the thirteen steps of a given sprinter. 

\subsubsection{Biomechanical features generation}

The estimated angles included the hip angle, defined as the angle between the trunk and the thigh, and the knee angle, defined as the angle between the thigh and the shank.
For each sprint, the running gait cycles are time-normalized, starting at 0\% (initial foot strike) and finishing at 100\% (next foot strike). Figure~\ref{fig:PP_UserStride} shows the evolution of the knee angle over time for 13 strides.
%

\section{VideoRun2D: Monomodal tracking results}\label{sec2}

Our first experiment presents the performance of the VideoRun2D system presented in Section~\ref{sec:VideoRunSystem}.
Table~\ref{tab:afterMonoResults} presents the results obtained for the angles of the hip and knee when we compared the trackers with the expert manual labeling, showing the mean RMSE values for the best angle of 5.30° for RTMPose-L. The results indicate that the RTMPose models are the most robust against errors examined in Section~\ref{sec:PP_TrackingModule}.

Our top three models are RTMPose-L, RTMPose-L*, and RTMPose-M, achieving RMSE values of 7.15°, 7.12°, and 6.99°, respectively. RTMPose-M was the model that performed the best in our experiments.
These results demonstrate the utility VideoRun2D Demo modules and the precision achieved by the considered trackers to the ground truth.

\begin{table}[t]
    \caption{RMSE values for all evaluated trackers
    (Section~\ref{sec:Tracking}). Boldface indicates the best
    result in each column.}
    \label{tab:afterMonoResults}
    \centering

    \begingroup
    \setlength{\aboverulesep}{0.15ex}
    \setlength{\belowrulesep}{0.15ex}
    \renewcommand{\arraystretch}{0.95}

    \resizebox{\columnwidth}{!}{%
        \begin{tabular}{lccccc}
            \toprule
            \multirow{2}{*}{Tracker}
            & \multicolumn{2}{c}{Hip angle~[°]}
            & \multicolumn{2}{c}{Knee angle~[°]}
            & \multirow{2}{*}{Mean~[°]} \\
            & Right & Left & Right & Left & \\
            \midrule
            ViTPose-B
            & 7.39 & 9.74 & 7.94 & 6.79 & 7.96 \\
            ViTPose-L
            & 7.00 & 9.53 & 5.75 & 6.54 & 7.21 \\
            RTMPose-M
            & \textbf{6.67} & \textbf{9.36}
            & 5.37 & 6.55 & \textbf{6.99} \\
            RTMPose-L
            & 6.94 & 9.75
            & \textbf{5.30} & 6.60 & 7.15 \\
            RTMPose-L$^*$
            & 7.01 & 9.73
            & 5.36 & \textbf{6.39} & 7.12 \\
            \bottomrule
        \end{tabular}%
    }

    \endgroup
\end{table}

\begin{table}[t]
    \caption{RMSE values obtained by averaging the pose estimates across
    trackers. Abbreviations: VL = ViTPose-L, RM = RTMPose-M,
    RL = RTMPose-L, and RL$^\ast$ = RTMPose-L$^\ast$
    (\(384 \times 288\) input). Boldface indicates the best result
    in each column.}
    \label{tab:bestFusion}
    \centering

    \begingroup
    \setlength{\aboverulesep}{0.15ex}
    \setlength{\belowrulesep}{0.15ex}
    \renewcommand{\arraystretch}{0.95}

    \resizebox{\columnwidth}{!}{%
        \begin{tabular}{lccccc}
            \toprule
            \multirow{2}{*}{Trackers}
            & \multicolumn{2}{c}{Hip angle~[°]}
            & \multicolumn{2}{c}{Knee angle~[°]}
            & \multirow{2}{*}{Mean~[°]} \\
            & Right & Left & Right & Left & \\
            \midrule

            VL + RM + RL + RL$^\ast$
            & 6.77 & 9.23 & \textbf{5.24} & 6.31 & 6.89 \\

            \midrule
            VL + RM + RL
            & 6.75 & 9.24 & 5.28 & 6.37 & 6.91 \\

            VL + RM + RL$^\ast$
            & 6.76 & \textbf{9.17} & 5.29
            & \textbf{6.29} & \textbf{6.88} \\

            VL + RL + RL$^\ast$
            & 6.87 & 9.36 & 5.27 & 6.32 & 6.96 \\

            RM + RL + RL$^\ast$
            & 6.77 & 9.29 & \textbf{5.24} & 6.33 & 6.91 \\

            \midrule
            VL + RM
            & \textbf{6.73} & 9.18 & 5.39 & 6.39 & 6.92 \\

            VL + RL
            & 6.88 & 9.44 & 5.35 & 6.43 & 7.03 \\

            VL + RL$^\ast$
            & 6.91 & 9.36 & 5.37 & 6.32 & 6.99 \\

            RM + RL
            & \textbf{6.73} & 9.34 & 5.26 & 6.44 & 6.94 \\

            RM + RL$^\ast$
            & 6.75 & 9.27 & 5.27 & 6.33 & 6.90 \\

            RL + RL$^\ast$
            & 6.92 & 9.52 & 5.26 & 6.36 & 7.01 \\
            \bottomrule
        \end{tabular}%
    }

    \endgroup
\end{table}

\begin{figure*}
    \centering
    \includegraphics[width=\textwidth]{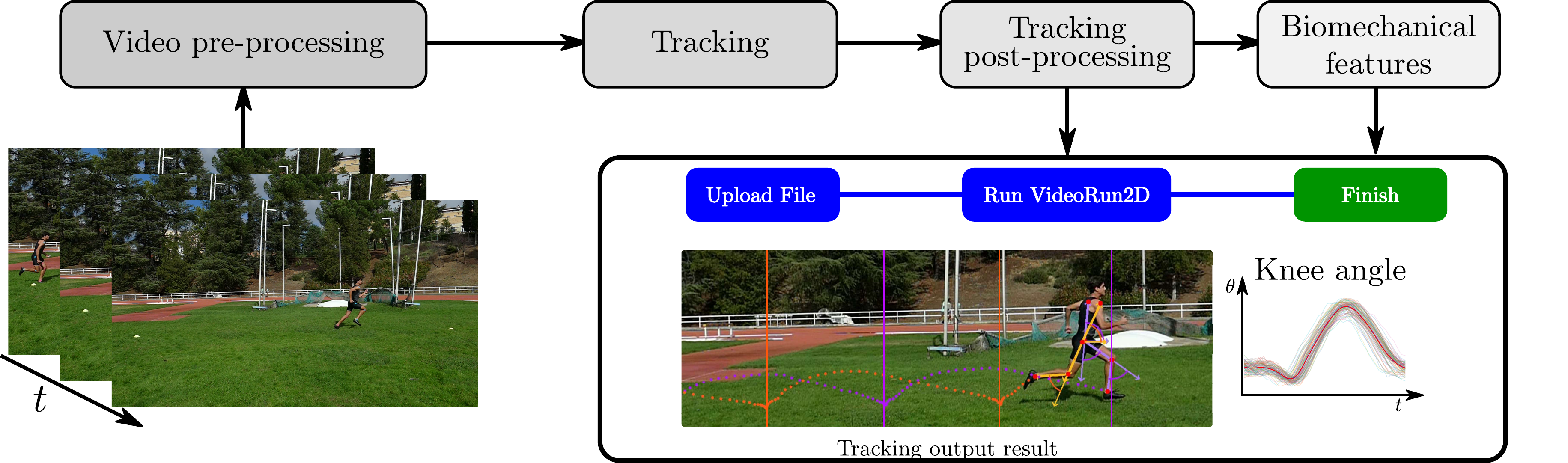}
    \caption{Block diagram of the VideoRun2D Demo platform operation.}
    \label{fig:Demo}
\end{figure*}

\begin{figure}
    \centering
    \includegraphics[width=\columnwidth]{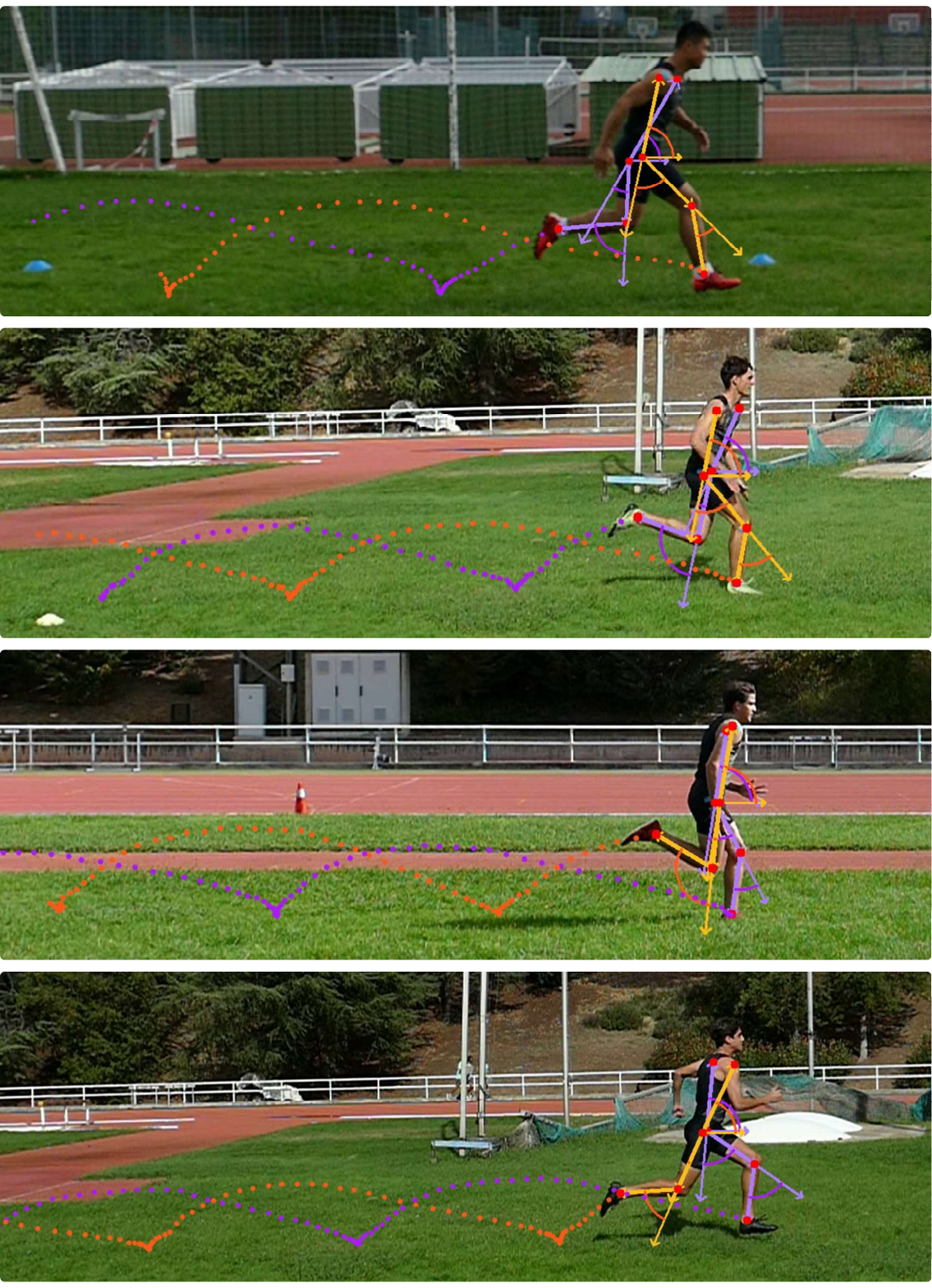}
    \caption{Graphical representation of joint points and angles calculated for four sprinters using the RTMPose model after the proposed tracking postprocessing. The orange lines and dots represent the right side of the corridor, while the purple lines and dots represent the left side. (Color image.)}
    \label{fig:Examples}
\end{figure}

\section{VideoRun2D: Multimodal tracking results}

We now experiment with a late-fusion strategy \cite{2018_INFFUS_MCSreview1_Fierrez} in which joint point predictions from the four best trackers are averaged to produce a final joint position.

Table~\ref{tab:bestFusion} shows the combination of trackers and shows a reduction in the RMSE for the different angles proposed. The combination of ViTPose-L, RTMPose-M, and RTMPose-L* produces the best average performance in predicting joint angles, with an RMSE of 6.88°, a decrease from the 6.99° initially obtained for the RTMPose-M tracker.

Specific angles, such as the right hip, do not show improvement when the trackers are combined, presenting good performance in a monomodal approach. 
On the other hand, the angle of the left knee shows an improvement, decreasing by 0.10° RMSE; the right knee shows an improvement, decreasing by 0.12° RMSE, and the angle of the left hip exhibits the greatest improvement, decreasing from 9.36° to 9.17° RMSE. 
This new model combination improves on the previous one
s in the literature. This improvement is attributed to the increased number of samples used, which enables a better generalization of the running problem from the sagittal plane by using more challenging scenarios and a new robust set of trackers.

\section{VideoRun2D: Demonstrator}

The main objective of this work is to promote our platform, where we make VideoRun2D Demo available to users.\footnote{\url{https://videorun2d.com/}} VideoRun2D Demo allows users to analyze video sprints with five different trackers, including ViTPose-B, ViTPose-L~\cite{xu2022vitpose}, RTMPose-M, and two versions of RTMPose-L~\cite{jiang2023rtmpose}. In addition, users can generate biomechanical analyzes of joint points and estimated angles.
This platform opens new opportunities for research in the sports community.

Figure~\ref{fig:Demo} shows the flow of the platform. Initially, the user must upload a sprint video. The user can aggregate the timestamps at which the sprinter appears. Then, the selected trackers infer the video and obtain the different time signals used for the angle estimation. Finally, the joint and angle signals can be shown in the right part of the platform, or we can generate a CSV with all the signals selected. Figure~\ref{fig:Examples} shows visual examples of the joint trajectories resulting.

\section{Discussion and Conclusion}\label{sec:Discussion}

This study presents VideoRun2D Demo, a framework for sprint analysis, and evaluates five state-of-the-art trackers by comparing their outputs against manual annotations on a dataset comprising 925 running gait cycles. Comparative experiments also assess the functionality of all modules in enhancing joint prediction across trackers, particularly in the presence of occlusions.

The estimated joint-angle curves (see Figure~\ref{fig:PP_UserStride}) are consistent with those reported in previous studies using marker-based trackers~\cite{ota2021verification, nagahara2017kinematics, yang2024improving} and Inertial Measurement Units (IMUs)~\cite{lin2023validity, rodrigues2020human}. Marker-based and IMU-based systems, however, require intrusive equipment and time-consuming data preparation. In contrast, the approaches presented in this work are non-invasive and cost-effective~\cite{
lin2023validity, yang2024improving}.

VideoRun2D Demo is competitive with marker-based solutions. However, further improvements are required for clinical applications, where an acceptable error range typically lies between 2° and 5°~\cite{mcginley2009reliability}.

VideoRun2D Demo provides a low-cost and non-invasive alternative for the biomechanical analysis of runners, with the potential to support future developments in individualized training and therapy. In addition, it may encourage broader participation from the scientific community in the advancement of biomechanical analysis tools.

In our future work, we will try to improve VideoRun2D by exploiting: 1) new image features based on visual attention \cite{dealcala25zoom}, and 2) frontier AI models such as VLMs \cite{dealcala2026demo2}. 

Analyzing and minimizing biases in core machine learning processes \cite{2023_ECAIw_LFIT-XAI_Tello} (e.g. gender and ethnicity biases \cite{serna27unravel}), detecting fake/synthetic manipulation in processed image/video streams \cite{paula2023gait,laura2026ava}, and respecting the related ongoing regulations around machine learning / AI systems \cite{irigoyen26risks} are also key research lines for us related to VideoRun2D.

\section{Acknowledgment}

Support by project M2RAI (PID2024-160053OB-I00, MICIU/FEDER) and Cátedra ENIA UAM-VERIDAS en IA Responsable (NextGenerationEU PRTR TSI-100927-2023-2). Work conducted within the ELLIS Unit Madrid. G. Garrido is supported by Banco Santander under a doctoral
grant. Thanks to Alejandro Carrillo for his work processing the data.

{
    \small
    \bibliographystyle{IEEEtran}
    \bibliography{main}
}



\end{document}